# Parameterized Synthetic Image Data Set for Fisheye Lens

Zhen Chen
Institute of Product and Process Innovation
Leuphana University Lüneburg
Lüneburg, Germany
zhen.chen@leuphana.de

Anthimos Georgiadis
Institute of Product and Process Innovation
Leuphana University Lüneburg
Lüneburg, Germany
georgiadis@uni.leuphana.de

*Abstract*—Based on different projection geometry, a fisheye image can be presented as a parameterized non-rectilinear image. Deep neural networks(DNN) is one of the solutions to extract parameters for fisheye image feature description. However, a large number of images are required for training a reasonable prediction model for DNN. In this paper, we propose to extend the scale of the training dataset using parameterized synthetic images. It effectively boosts the diversity of images and avoids the data scale limitation. To simulate different viewing angles and distances, we adopt controllable parameterized projection processes on transformation. The reliability of the proposed method is proved by testing images captured by our fisheye camera. The synthetic dataset is the first dataset that is able to extend to a big scale labeled fisheye image dataset. It is accessible via: http://www2.leuphana.de/misl/fisheye-data-set/.

## I. Introduction

Machine vision technology become a powerful tool on the field of surveillance, automated inspection, robot control and navigation[1]. In the task of robot navigation, a navigation system builds up a map and localizes the position by continually updating the robust landmarks from the vision system. According to comparing landmarks stored in the memory, the navigation system recalls scenarios that passed by and then determines the location of the robot on an excited map. The mentioned map building and position searching procedure are called simultaneous localization and mapping (SLAM) in the machine vision community[2].

The fisheye camera, benefited from the very large field of view (FOV), provides rich vision information in SLAM. The lens from the camera is capable of achieving over 180 degrees' visual angles instead of the general 40 to 60 degrees. Therefore, the fisheye lens achieves more and more attention in the application of visual navigation systems. D. Caruso, et al. apply a fisheye cameras to a large-scale visual SLAM and develop the monocular SLAM algorithm[3]. M. Bertozzi, et al. develop a system for pedestrian and vehicle detection and tracking by means of fisheye images[4].

Deep convolution neural networks (DCNN) is another popular research field in the machine vision community. Benefits from enormous computational capabilities of GPU, the DCNN is trained deeper and deeper. The recent research shows that deeper convolution neural networks are able to describe more complex features hidden behind the image[5]. Some research results on the application of the DCNN are available. D. Zhe, et al. use the DCNN to evaluate the quality of images with the view of the human on the classes of good and bad[6]. Y. Kawano, et al. improve food recognition accuracy by using the feature abstracted from the DCNN[7].

The DCNN is a potential method to explore the hidden meaning behind the fisheye image. The main challenge on machine learning engineering is how to train a generalize neural network model without overfitting. Some skills are helpful in preventing overfitting, such as parameter regularization and dropout. However, the most effective way for solving such problem is to enlarge the size of the training dataset. Unfortunately, collecting and labeling fisheye images are extremely time-consuming. So it is not feasible to train a neural network detector on novel categories. In this paper, we propose to bypass the expensive collecting procedure and build a synthetic training image dataset with an existing perspective image dataset. For the parameterized synthetic image dataset, the optical calibration is not necessary, since the camera's parameters are pre-defined on the process of the synthesis. With the synthesis fisheye dataset, verification and evaluation of a new algorithm can be completed with a virtual optical environment. Under the following sections, we analyze the geometrical model of the fisheye lens, and introduce the algorithm and the procedures for generating synthetic fisheye images. The validation the proposed method is tested by new neural network detector. The image dataset is accessible via: http://www2.leuphana.de/misl/fisheye-data-set/

## II. The Structure of the Fisheye Lens

The fisheye camera system is a typical non-linear system, in which the spatial resolution decreases from the image's center to the edge. Considering a series of factors, like size, focus, iris and geometry, different camera manufacturers have different fisheye system designs, especially fisheye lenses [8]. A typical fisheye lens is consisting of several layer of highly-integrated optical glasses, such as convex lens, concave lens and filter lens.

When complex glasses combination is applied for a unique light route design, it is difficult to describe a fisheye geometry with a single model. Thus, several standard imaging models are raised based on the typical fisheye projection model. As shown in Fig. 1, the ray's incidence angle φ and the corresponded reflection angle β are different because of the light refraction. With different expression between the image

radius r and the incidence angle $\varphi$, the fisheye projection models are classified as[9]:

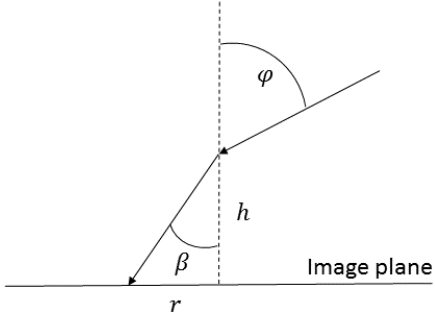

Figure 1.  Fisheye projection geometry.

Equidistant projection:
$$r = h \cdot \varphi \quad (1)$$
Stereographic projection:
$$r = 2h \cdot \tan\frac{\varphi}{2} \quad (2)$$
Equisolid-angle projection:
$$r = 2h \cdot \sin\frac{\varphi}{2} \quad (3)$$
Orthographic projection:
$$r = c \cdot \sin\varphi \quad (4)$$

To project a real object point into a fisheye image, the camera coordinate system $(x, y, z)$ and the image coordinate system $(x', y')$ is defined. The projection model for the above projection could be expressed as follows[10]:

Equidistant projection:
$$x' = c \cdot \frac{\arctan\frac{\sqrt{x^2+y^2}}{z}}{\sqrt{\left(\frac{y}{x}\right)^2+1}} + x'_0 + \Delta x' \quad (5)$$

$$y' = c \cdot \frac{\arctan\frac{\sqrt{x^2+y^2}}{z}}{\sqrt{\left(\frac{y}{x}\right)^2+1}} + y'_0 + \Delta y' \quad (6)$$

Stereographic projection:
$$x' = c \cdot \frac{\tan\left(\frac{1}{2} \cdot \arctan\frac{\sqrt{x^2+y^2}}{z}\right)}{\sqrt{\left(\frac{y}{x}\right)^2+1}} + x'_0 + \Delta x' \quad (7)$$

$$y' = c \cdot \frac{\tan\left(\frac{1}{2} \cdot \arctan\frac{\sqrt{x^2+y^2}}{z}\right)}{\sqrt{\left(\frac{y}{x}\right)^2+1}} + y'_0 + \Delta y' \quad (8)$$

Equisolid-angle projection:

$$x' = c \cdot \frac{\sin\left(\frac{1}{2} \cdot \arctan\frac{\sqrt{x^2+y^2}}{z}\right)}{\sqrt{\left(\frac{y}{x}\right)^2+1}} + x'_0 + \Delta x' \quad (9)$$

$$y' = c \cdot \frac{\sin\left(\frac{1}{2} \cdot \arctan\frac{\sqrt{x^2+y^2}}{z}\right)}{\sqrt{\left(\frac{y}{x}\right)^2+1}} + y'_0 + \Delta y' \quad (10)$$

Orthographic projection:

$$x' = c \cdot \frac{\sin\left(\arctan\frac{\sqrt{x^2+y^2}}{z}\right)}{\sqrt{\left(\frac{y}{x}\right)^2+1}} + x'_0 + \Delta x' \quad (11)$$

$$y' = c \cdot \frac{\sin\left(\arctan\frac{\sqrt{x^2+y^2}}{z}\right)}{\sqrt{\left(\frac{y}{x}\right)^2+1}} + y'_0 + \Delta y' \quad (12)$$

In which $(x'_0, y'_0)$ are principle point. The correction parameters $\Delta x, \Delta y$ are defined as:
$$\Delta x = x' \cdot (A_1 c + A_2 r'^4 + A_3 r'^6) + B_1 \cdot (r'^2 + 2x'^2) + 2B_1 x' y' + C_1 \cdot x' + C_2 \cdot y' \quad (13)$$

$$\Delta y = y' \cdot (A_1 r'^2 + A_2 r'^4 + A_3 r'^6) + 2B_1 x' y' + B_2 (r'^2 + 2y'^2) \quad (14)$$

In which $(A_1, A_2, A_3)$ are radial distortion parameters, $(B_1, B_2)$ are decentering distortion parameters and $(C_1, C_2)$ are the horizontal scale factor and the shear factor.

### III. SYNTHETIC FISHEYE IMAGE

Perspective images captured by the traditional camera are comparatively easier to be collected. Even though, it takes much effort to manually label each image, especially when the number of images is large. For a new algorithm verification, labeling images occupies a large amount of time. In this paper, we propose to synthesize a fisheye image dataset by transforming a well-known labeled perspective image dataset. By using the proposed method, we are able to prepare a useful training dataset quickly. The developed synthetic fisheye image dataset is accessible online, to my knowledge, which is the first public synthetic fisheye image dataset by transforming a known perspective image dataset.

There are several large-scale labeled image datasets on the natural scene. The popular datasets in the machine vision community are MNIST, CIFAR10 / CIFAR100, Imagenet and so on, which are often used for training and testing a neural network (NN). The MNIST dataset is a size-normalized hand written digits, which is often used by beginners for practicing[11]. CIFAR10 and CIFAR100 dataset, collected by K. Alex, N. Vinod and H. Geoffrey, have maximum 80 million color images in 100 classes. The resolution of the images is unified to $32 \times 32$ for each class[12]. In the work from this paper, we choose Imagenet Large Scale Visual Recognition Challenge (ILSVRC) 2012[13] training dataset as the raw image dataset, which provides 1.2 million labeled images in 1000 categories. One reason we choose this dataset is that high-resolution raw images maintain an amount of detail information during the synthesis. Another reason is that the scale of the synthetic dataset is flexible because it can be extended to 1000 classes. To confirm the reliability of the proposed method, we select 12 classes' images from indoor objects and transform them into fisheye images. These 12 classes consist of ballpoint pen, cellular telephone, desktop computer, espresso maker, printer, projector, shopping cart, stone wall, television, wall clock, wardrobe and water bottle.

For the synthesis, we expect to create the effects like observing from different distances and angles. Based on this idea, an original image from the ILSVRC can be transformed into dozens of fisheye images, especially by adding center bias into the transformation. Besides, we hope to keep the same visible regions between the raw image and the synthetic fisheye image. To realize that, we unify the image's size before the transformation: an images' longer side is first scaled down to 512 pixels and then extend the new images' shorter side to 512 pixels by filling black pixels.

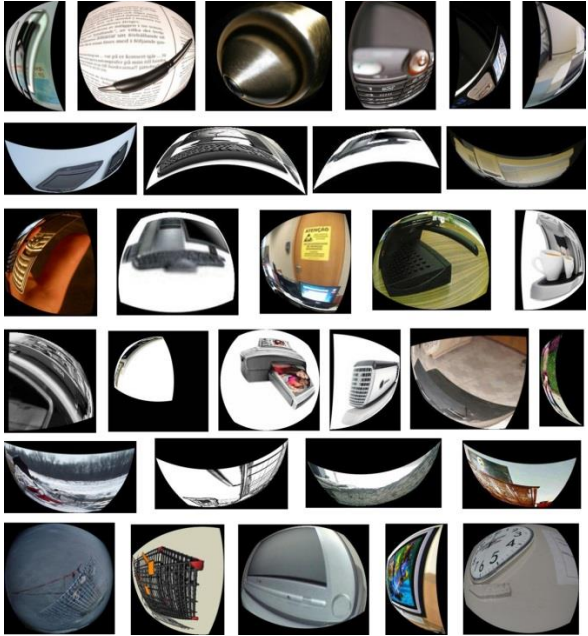

Figure 2. Representative samples of the synthetic fisheye images

Equidistant projection is applied on the transformation in this paper, as shown on equation (5) and (6). To obtain the synthetic effects for different viewing angles and distances, the center offset for the synthetic images is set in which the distance ranges from 0 to 1/6 of the image's width and the degree ranges from 26 to 35 degrees separately. Representative samples of the synthetic fisheye images can be seen in the Fig. 2.

## IV. EXPERIMENTAL TEST

### A. The structure of the neural network

Convolutional Neural Networks (CNN) is composed of a succession of convolutional layers and fully connected layers. Each layer receives feedback only from the previously connected layer. Therefore, it is time-consuming to train an entire convolutional network from scratch on a large-scale dataset. Fine-tuning a pre-trained NN is a way to accelerate the parameter's update for a special task[14]. On the fine-tuning, the preserved parameters from a pre-trained NN would be transferred to a new NN.

The neural network architecture we use is Alexnet, which wins the ImageNet Large Scale Visual Recognition Challenge (ILSVRC) in 2012[15]. The Alexnet consists of five convolutional layers, three fully-connected layers and a 1000 classes softmax layer. The acceleration convergence on the convolutional and fully connected layer is applied by the ReLU non-linearity, instead of the sigmoid and the tahn function. The pooling layers are inserted at the first, the second and the fifth layer. In the pooling layers, an overlapping pooling method is utilized in order to eliminate overfitting problem. The architecture of Alexnet is shown in Fig. 3.

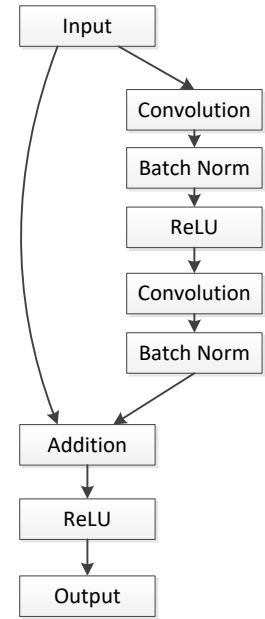

Figure 3. The neural network architecture for Alexnet

In this paper, we test the synthetic images in 12 classes, for which the original softmax layer is replaced by a new 12 classes softmax classifier. Then, the parameters of the pre-trained NN are transferred to a new model. Considering the great difference between the dataset of the pre-trained NN and the new synthetic fisheye dataset, we also fine-tune the lower layer of the neural network to ensure the adjustment of the low level feature.

## B. DCNN training

In this section, we practice the knowledge transferring from a pre-trained Alexnet model to a new fisheye model in order to confirm the reliability of the proposed method. Besides, we test the new model's accuracy by images captured by a fisheye camera. The pre-trained model is trained from ILSVRC 2012 image dataset.

The tests run on Dell Precision T5500 workstation that configures dual Intel Xeon Processor E5645 (6 cores, 2.4 GHz) and MSI Geforce GTX 1080 Graphic card (2560 CUDA cores, 8 GB RAM). The implementation is based on the Deep Learning Library Caffe. The synthetic images are divided into three groups: 100349 sample images to train the neural network, 33452 sample images to validate the neural and 33449 sample images to test the accuracy of the neural network. We give the softmax layer a larger learning rate to accelerate the learning process. The softmax layer is fine-tuned with the setting: Stochastic gradient descent (SGD) sets as 0.01 base learning rates and 0.5 times step down for every 20 percentage step size. The accuracy of the fine-tuned model on the test dataset reaches to 82.5%.

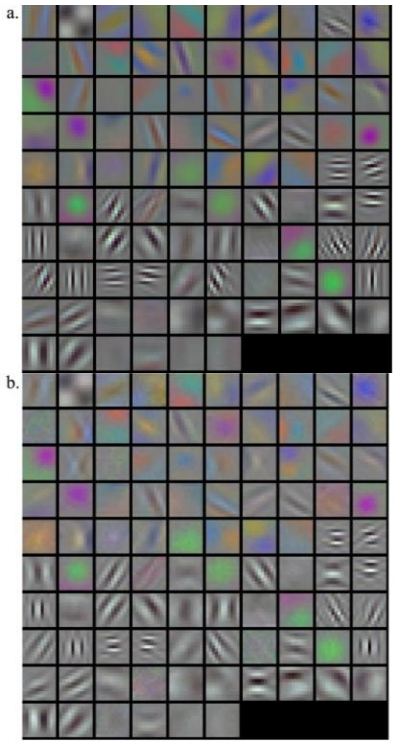

Figure 4. The visualization of the detector on the first layer: a. perspective image neural network model, b. fisheye image neural network model

The fine-tuned model has a 12 class's softmax classifier which is supported by 12 neurons for every class. We also fine-tune the pre-trained model with 12 classes perspective image on the same object, in order to reveal the changes between perspective image classification model and fisheye image classification model. Fig. 4 illustrates the filters weight on the first CONV layer for perspective NN model and fisheye NN model respectively. From the visualization of the weight, fisheye NN model has similar filters weights as perspective NN model. But the weight image of the fisheye NN model has lower image sharpness, which weakens the expression of sharp edges and corners in the image. The changes on the first layer demonstrate the improvement of the new model for the curved images.

## C. Image classification test

In the experiments, fisheye images are collected using Fujifilm Fujinon F-FE185C057HA-1 lens (0.1 m to infinity focusing range, 185-degree viewing angle on 2/3 inch sensor) and Point Grey Grasshopper 3 camera (GS3-PGE-50S5C-C, 2448 x 2048 resolution, 5.0 MP megapixels, 2/3 inch sensor). Six objects (two ballpoint pens, three computers and a plastic bottle) are placed on the desk. Images are shot from three distance ranges separately: short distance (0.3-0.8 m), middle distance (0.8 - 2 m) and long distance (2 - 5 m).

With the increasing of the distance between the object and the camera, the number of pixels occupied by each object decreases. The object that close to the edge of the image occupies fewer pixels because of the high compression ratio at the edge of the lens. We collect three images for each distance ranges. Considering external affection on the experiments, we parallel place two ballpoint pens in the distance of 5 cm and also set three computers closely. The resolution ratio of a raw fisheye image is 1800 x 1800 pixels.

Five groups of objects' images are collected from different distances and angles, in which two pens' images are collected in same images and three computers are together. Interest regions are artificially selected and cropped. To fit the size of the input image for neural network, we scale down the images the longer side to 350 pixels, as shown in Fig. 5.

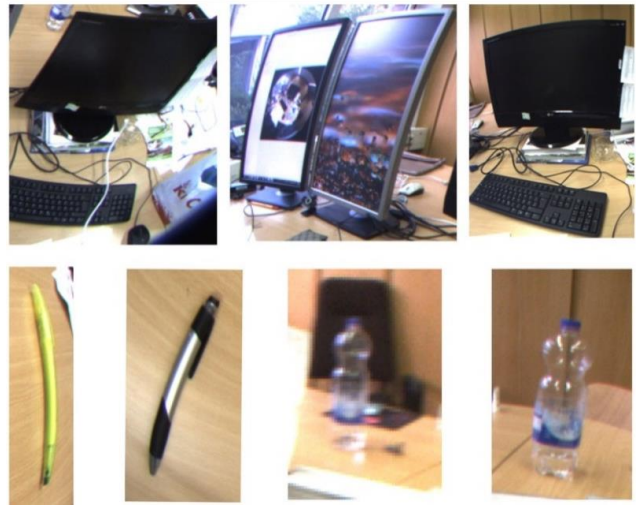

Figure 5. The images captured by fisheye camera.

We test the performance of the new DCNN mode on the acquired photos. The assessment of the DCNN classification performance is by its top-1 and top-3 prediction reliability. As shown in the Fig. 6, the confidences levels of the top-1 classification for each image are in the range of 75% and 95%, except one test for a computer image attends to 48.1%. The overall accuracy for the top-1 prediction on the collected images attends to 97.6%.

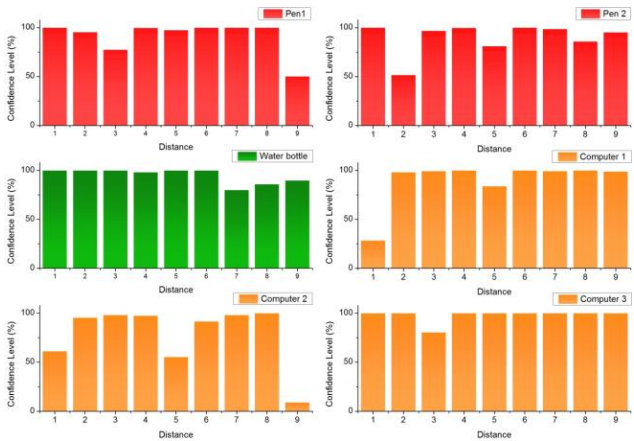

Figure 6.  Top-1 classification prediction for the captured fisheye images

## V. CONCLUTION

DCNN is widely used in the intelligent system, especially on the field of machine vision. For a large scale of DCNN training, the absence of the labeled images become the biggest disadvantage. In this paper, we propose to synthesize the fisheye image by transferring the labeled perspective image in order to reuse the existing resources. The transformation applies the equidistant projection. The proposed method is approved by the trained fisheye NN model and is tested on fisheye images captured by real fisheye camera. The results show that the synthetic fisheye image could be used on the big scale DCNN training. The synthetic fisheye dataset is available online, which is the first labeled fisheye dataset created by using existing dataset.